\documentclass[10pt,journal,compsoc]{IEEEtran}
\ifCLASSOPTIONcompsoc
  \usepackage[nocompress]{cite}
\else
  \usepackage{cite}
\fi
\ifCLASSINFOpdf
\else
\fi
\usepackage{CJKutf8}
\usepackage{graphicx}
\usepackage{subfigure}

\usepackage{url}

\usepackage{amsmath,amssymb,amsfonts}
\usepackage{algorithmic}
\usepackage{algorithm}

\usepackage{booktabs}
\usepackage{multirow}

\usepackage{microtype}
\usepackage{ragged2e}

\usepackage{pifont}
\usepackage{booktabs}
\usepackage{makecell}

\usepackage[T1]{fontenc}
\usepackage[most]{tcolorbox}
\newtcolorbox{promptbox}[2][]{
	enhanced,                
	breakable,               
	colback=gray!5,          
	colframe=gray!50,        
	title={#2},              
	fonttitle=\bfseries\sffamily, 
	coltitle=black,          
	attach boxed title to top left={yshift=-2mm, xshift=2mm}, 
	boxed title style={
		colback=gray!15,       
		colframe=gray!50,      
		sharp corners,         
		boxrule=0.5pt          
	},
	fontupper=\small\ttfamily, 
	lines before break=2,      
	#1                       
}

\begin{document}
\begin{CJK*}{UTF8}{gbsn}
%
\title{Sledgehammer or Scalpel? A Fine-grained Adaptive Framework for Implicit Hate Speech Detection}

\author{Han~Wang,
	Hanqing~Tao,
	Xuesong~Wang,
	and~Yi~Zhu%
	\thanks{H. Wang, Q. Tao, and X. Wang are with the School of
		Information and Control Engineering, China University of Mining
		and Technology, Xuzhou 221100, China
		(e-mail: wanghanhan0102@163.com; hqtao@cumt.edu.cn; wangxuesongcumt@163.com).}%
	\thanks{Y. Zhu is with the School of Information Engineering,
		Yangzhou University, Yangzhou 225127, China
		(e-mail: zhuyi@yzu.edu.cn).}%
	\thanks{Corresponding author: Yi Zhu.}%
}
\markboth{Journal of \LaTeX\ Class Files,~Vol.~14, No.~8, August~2026}%
{Shell \MakeLowercase{\textit{et al.}}: Bare Demo of IEEEtran.cls for Computer Society Journals}
%



\IEEEtitleabstractindextext{%
\begin{abstract}
	\justifying
Unlike explicit attacks with obvious profanity, implicit hate speech hides malice within seemingly compliant expressions through metaphors and contextual hints, making its detection in online content review challenging. Many PLM- or LLM-based methods apply the same processing pipeline across samples, which may overlook differences in information requirements and introduce unnecessary computation. We observe that online hate speech is not monolithic but manifests in varied forms. We therefore define three fine-grained categories: Shallow, Targeted, and Context-Dependent. Accordingly, we propose \textbf{F}ine-grained \textbf{A}daptive \textbf{I}mplicit Hate speech \textbf{D}etection (FAID), a novel framework that first performs fine-grained classification and then adapts to specific categories. Specifically, for Shallow samples with surface-identifiable intents, the framework adopts lightweight Prompt-Tuning for rapid classification; for Targeted comments that bind malicious intent to concealed targets, we design knowledge augmentation to iteratively refine the model and reveal hidden targets; for Context-Dependent comments lacking background information, we utilize an agentic framework that automatically generates prompts to evolve context, infer missing background information and identify ambiguous malicious intents. This adaptive architecture focuses computational resources on complex implicit samples while avoiding redundant reasoning for shallow samples. Experiments on four benchmark datasets demonstrate that FAID significantly outperforms SOTA baselines.
\end{abstract}

}

\maketitle

\IEEEdisplaynontitleabstractindextext

%
\IEEEpeerreviewmaketitle

\section{Introduction}
\fbox{\begin{minipage}{8.2cm}
		\textbf{Content Warning:} The paper contains content that some may find disturbing or offensive, including content that is discrimi native, hateful, or violent in nature.
\end{minipage}}
\vspace{0.6em}
\IEEEPARstart{T}{he} exponential growth of the Internet and social media has promoted the popularization of online expression, allowing users to express various types of opinions. However, this has gradually turned online communities into a breeding ground for hate speech, which can evolve over time into serious consequences like cyberbullying or even cyber violence \cite{anjum2024hate}. Although almost all online platforms have deployed algorithms to curb explicit hate speech, which contains obvious malicious words and direct attacks, these methods are often ineffective against the more covert implicit hate speech \cite{kim2022generalizable}. This is because implicit hate speech uses techniques like sarcasm and metaphor to convey malicious intent without using profanity. Therefore, in practice, achieving rapid and accurate detection of such content, which covers a wide variety of complex expressions, faces severe challenges \cite{chen2024survey}.

\begin{table*}[h]
	\centering
	\caption{Examples of Different Implicit Hate Speech Categories}
	\label{tab:hate_examples}
	\begin{tabular}{c l c}
		\toprule
		\textbf{No.} & \textbf{Example} & \textbf{Categories} \\
		\midrule
		1 & Black people should just go back to the cotton fields. & Shallow \\
		2 & The Atlantic is actually a big cup of bubble tea. & Targeted\\
		3 & Look at how `brave' she is for wearing that. & Context-Dependent \\
		\bottomrule
	\end{tabular}
\end{table*}

Most existing implicit hate speech detection methods adopt uniform processing paradigms, ignoring the great heterogeneity within implicit hate speech. As shown in Table~\ref{tab:hate_examples}, for Sample 1, it directly links ``black people'' to well-known racial discrimination. Its hateful intent has long become public knowledge, allowing shallow classifiers to efficiently handle the detection task.
Sample 2 alludes to historical events such as enslaved people being thrown overboard during the triangular trade, or capitalists dumping milk during the Great Depression. The model needs rich external knowledge to translate seemingly non-malicious or even nonsensical statements into the referred events and map the attacked object to a specific group. For Sample 3, where the word ``brave'' seemingly expresses praise literally but may be satirizing someone's body or inappropriate dressing in a specific context or trending meme, the model needs deep contextual reasoning to see through the sarcastic intent behind ``brave.'' This heterogeneity leads to a serious problem that deploying complex reasoning models for simple samples causes computational redundancy and wastes resources, while using shallow classifiers for samples that require understanding internal references and contexts leads to detection failure. Crucially, mixing the latter two categories exacerbates the detection bottleneck. The second category mainly relies on factual knowledge retrieval, while the third relies on logical deduction. If a pure reasoning model without external knowledge is used for the second category, it often leads to factual hallucinations; whereas forcing knowledge retrieval on the third category may introduce irrelevant noise, masking the sarcastic intent. Based on the observation of these real events, in order to address the detection bottleneck problem caused by their mixed handling, we define implicit hate speech into three fine-grained categories, consisting of Shallow, Targeted, and Context-Dependent categories.

Facing the three distinct categories mentioned above, existing methods based on fine-tuning Pre-trained Language Models (PLMs) \cite{kim-etal-2023-conprompt,caselli2020hatebert} or reasoning with Large Language Models (LLMs) \cite{jafari2024target,yang2023hare} expose clear limitations. Although PLMs are highly efficient at capturing explicit features, they often lack relevant knowledge, making it difficult for them to understand metaphors and sarcasm like Targeted and Context-Dependent categories. In contrast, LLMs excel at understanding complex metaphors and contexts, and have the potential to handle Targeted and Context-Dependent comments. However, relying solely on LLMs for complex tasks often leads to severe hallucinations, which many current approaches try to mitigate using multi-agent frameworks \cite{sun2026rethinking} or chain-of-thought \cite{wei2022chain}. Unfortunately, LLMs still exhibit limitations in sarcasm understanding \cite{zhang2025sarcasmbench}. Moreover, these methods can introduce substantial computational overhead, making it difficult to meet the real-time detection requirements of social media. The trade-off between detection performance and computational cost has also received attention in recent research \cite{gui2026eamkd}.

When addressing implicit hate speech with different information requirements, should we wield a straightforward but imprecise ``Sledgehammer'', or a meticulous but time-consuming ``Scalpel''? To avoid the mismatch with data features caused by forcing a single tool to fit all scenarios, we abandon static and single processing methods and adopt a \textbf{F}ine-grained \textbf{A}daptive \textbf{I}mplicit Hate speech \textbf{D}etection (FAID) method. Specifically, we decouple the input data into the three granularities mentioned above and then dynamically route them to specialized processing modules. We deploy the straightforward ``Sledgehammer'' of rapid classification by lightweight Prompt-Tuning for easily identifiable shallow samples, while wielding the meticulous ``Scalpel'' for more covert malice: a knowledge-augmented iterative model for targeted attacks, and the Agentic Context Engineering (ACE) \cite{zhang2025agenticcontextengineeringevolving} framework for context-dependent samples. The framework selects detection mechanisms based on the information requirements of samples, balancing detection performance and processing cost while addressing the needs of different samples. The main contributions of this paper are summarized as follows:

\begin{itemize} 
	\item We propose a fine-grained taxonomy for implicit hate speech detection, defining three categories: Shallow, Targeted, and Context-Dependent. This categorization is based on the information required for judgment, distinguishing reliance on surface-level semantics, interpretation of hidden targets, and inference of missing context.
	\item We develop FAID to connect these information requirements with different detection mechanisms. Based on the predicted category, the router directs each comment to prompt-based classification, knowledge-augmented iterative learning, or agentic context reasoning, aligning the reasoning strategy and computational cost with the requirements of each category.
	\item Extensive experiments on four benchmark datasets show that our framework achieves State-of-the-Art (SOTA) performance compared to existing baselines. Furthermore, the average inference time is significantly lower than existing LLM-based methods, achieving the best balance between detection performance and computational efficiency.
\end{itemize}

Our source code and fine-grained annotated datasets are publicly available at \url{https://github.com/HanWang0102/FAID}.

\section{Related Work}
\subsection{Implicit Hate Speech Detection}
Implicit hate speech conveys hostile intent indirectly. Its detection therefore requires models to recover latent meaning, attack targets, or implicit rationales rather than relying only on surface lexical cues \cite{mathew2021hatexplain}. Existing methods are mainly based on PLMs or LLMs. PLM-based methods enhance textual representations through domain-adaptive pretraining and contrastive learning, achieving strong classification performance at relatively low computational cost \cite{kim2022generalizable,kim-etal-2023-conprompt,caselli2020hatebert}. However, they remain limited in identifying concealed targets and missing context. LLM-based methods improve the interpretability and detection of complex samples through explanation generation, target identification, and explicit reasoning \cite{yang2023hare,zhang-etal-2024-dont-go,nirmal-etal-2024-towards,jafari2024target}, but remain prone to hallucinations and incur high computational costs. 

The detection requirements of different samples vary not only in reasoning difficulty but also in the information needed for judgment. Therefore, when selecting a detection mechanism, it is necessary to consider whether it matches the needs of each sample.

\subsection{Prompt Engineering} 
Prompt engineering adapts pretrained models to downstream tasks through the design of input templates or task instructions \cite{liu2023pre}.  For masked PLMs, prompt engineering usually adopts a cloze-style template learning paradigm, which bridges the gap between pre-training objectives and downstream tasks by predicting the [MASK] token in the template \cite{kan2023composable,zhu2024prompt}. This paradigm mainly relies on two core components: the template \cite{schick2021exploiting, ZHU2024123248} and the verbalizer \cite{gao2020making,hu2022knowledgeable, ZHU2025109589}. For generative LLMs, prompt engineering mainly guides models to perform reasoning and generation through natural language instructions, and makes model outputs more logical through in-context learning \cite{Brown2020LanguageMA}, chain-of-thought reasoning \cite{wei2022chain}, iterative optimization based on feedback \cite{shinn2023reflexion}, and structured reasoning \cite{yue2026interarm}. Beyond individual prompts, ACE organizes contextual information as evolving playbooks, accumulating and refining strategies through structured incremental updates \cite{zhang2025agenticcontextengineeringevolving}. 

These techniques support different processing mechanisms for classification, explanation generation, and contextual reasoning. Adaptive routing can coordinate their use according to the information requirements of individual samples.

\subsection{Adaptive Routing}
Adaptive routing selects computation paths based on input features, allocating computational resources according to the processing requirements of different samples. In computer vision, HydraNets select a subset of specialized branches for each image \cite{mullapudi2018hydranets}, while Dynamic Routing Networks use lightweight routing networks to select the required transformation branches, producing computation paths that vary with the input \cite{cai2021dynamic}. Beyond network branch selection, CARE-Edit assigns tokens to experts with different specializations based on multimodal conditions and diffusion timesteps \cite{wang2026careedit}. In language tasks, RoutingGen selects between few-shot prompting and structured reasoning according to the difficulty of code-generation tasks \cite{li2026routinggen}. These studies show that dynamically selecting computation paths, experts, or reasoning strategies based on the input has been implemented in various ways.

In implicit hate speech detection, DuPL uses a Confusion Judge to determine whether a comment requires further argumentation about latent hate components \cite{sun2026rethinking}. Based on the initially mined components, it selectively sends difficult comments with ambiguous initial judgments to the subsequent argumentation stage. However, deciding whether a comment requires further argumentation does not directly identify the type of information it lacks. Comments that require hidden targets to be resolved or missing context to be inferred may therefore enter the same argumentation process. Their information requirements are not explicitly matched to different processing mechanisms. Our FAID focuses on differences in the information requirements of implicit expressions, as resolving hidden targets requires target explanations and knowledge augmentation, whereas inferring missing context requires contextual reasoning. This design uses the information requirements of samples as the basis for category routing and matches each category with lightweight classification, knowledge-augmented iterative learning, or agentic context reasoning to select an analysis method suited to each sample.

\section{Methodology}
\subsection{Problem Definition and Framework Overview}

\renewcommand{\dblfloatpagefraction}{0.7}
\begin{figure*}[htbp]
	\centering
	\includegraphics[scale=0.95]{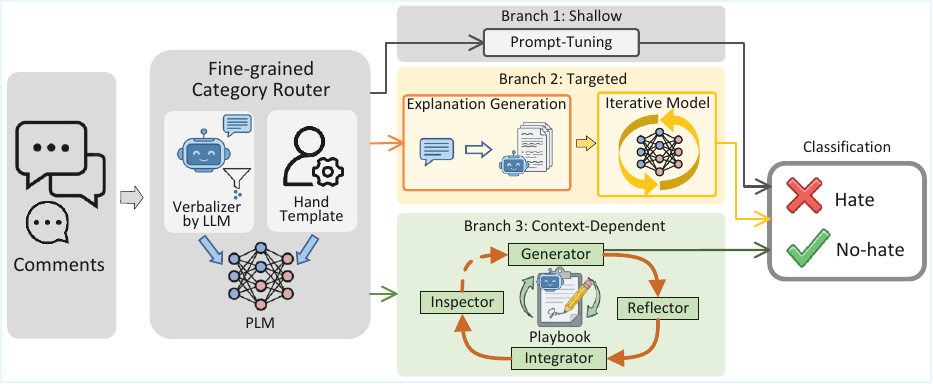}
	\caption{The overall framework of the proposed FAID. The input comments are first routed into three fine-grained categories (Shallow, Targeted, and Context-Dependent) via a Fine-grained Category Router module. While Shallow is classified by Prompt-Tuning, Targeted comments undergo knowledge augmentation via LLM-generated explanations and are processed by an iterative model to capture latent semantics. Context-Dependent comments are handled by an enhanced ACE framework, which utilizes a ``Generator-Reflector-Integrator-Inspector'' multi-agent loop to dynamically evolve a playbook for inferring missing contexts and ensuring accurate detection.}
	\label{framework}
\end{figure*}
The implicit hate speech detection task is formulated as a typical short-text classification problem. For an input comment $x$, the goal is to predict its label $y \in \{\text{hate}, \text{no-hate}\}$. We observe that different types of comments have distinct modes of expression and reasoning complexities, some display easily identifiable surface-level semantics, some imply attack targets through references, and others require contextual clues to be identified. Feeding all comments into a single model may lead to feature confusion and performance degradation. Therefore, we propose a Fine-grained Adaptive Implicit Hate speech Detection framework. 

As illustrated in Figure~\ref{framework}, the original comments are first automatically divided into three categories through fine-grained categorization, and dedicated heuristic strategies are then designed to address the linguistic features of each category. For shallow comments, we directly use the same lightweight Prompt-Tuning method from the fine-grained classification stage to perform the binary classification; for Targeted comments, we guide an LLM to analyze the comment to uncover specific references to concatenate this analysis with the original comment, and input it into an iterative model to better capture latent semantics; for Context-Dependent comments, drawing on the prompt evolution idea from the ACE method, we construct a ``Generator-Reflector-Integrator-Inspector`` cyclic agent to dynamically evolve a Playbook, which guides the LLM to complete detection after inferring the appropriate context. Unlike traditional implicit hate speech detection, we do not use a single strategy to cover all cases. Instead, we improve model performance on complex categories through category-adaptive heuristic design.

\subsection{Fine-grained Category Router \label{method_to_three}}

\begin{figure*}[htbp]
	\centering
	\includegraphics[scale=1.2]{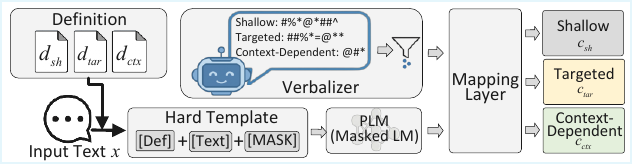}
	\caption{Framework of fine-grained categorization using Prompt-Tuning, corresponding to the gray shaded area in Figure~\ref{framework}. The input comment is inserted into a handcrafted hard template, passed through the PLM embedding and transformer layers, and processed by MaskLM. The verbalizer then maps predicted label words to predefined categories, and the classifier determines the final fine-grained category.}
	\label{prompt-tuning}
\end{figure*}

As illustrated in Figure~\ref{prompt-tuning}, the category router classifies each comment according to its information requirements and selects the corresponding detection branch. Formally, for an input comment $x$, the router predicts a category $c \in C$, while the selected branch produces the final no-hate/hate label. The category set is defined as follows:
\begin{equation}
	C = \{c_{sh}, c_{tar}, c_{ctx}\} 
	\label{defineC}
\end{equation}
where $c_{sh}$ refers to shallow comments, $c_{tar}$ refers to Targeted comments, and $c_{ctx}$ refers to Context-Dependent comments. Since prompt-tuning can achieve strong performance without adjusting model parameters, we apply Prompt-Tuning to perform this categorization. Prompt-tuning reformulates the task as a masked language modeling problem, where the input comment is placed into a handcrafted template ($T(\cdot)$), and the PLM predicts the word at the masked position. By solving this cloze-style problem, the model infers the fine-grained category $c \in C$ of the comment.

\subsubsection{Design of Handcrafted Templates}

In this Prompt-Tuning approach, the template plays an important role. We design handcrafted hard templates that the classification task require establishing clear classification boundaries. First, we define the three categories $D = \{d_{sh},d_{tar},d_{ctx}\}$ as follows:

\begin{align}
	d_{sh} &= \text{``Comments with easily identifiable surface-level } \notag \\
	&\quad \text{intents''} \\[6pt]
	d_{tar} &= \text{``Comments with identifiable time, place, person,} \notag \\
	&\quad \text{or event clues''} \\[6pt]
	d_{ctx} &= \text{``Comments with ambiguous meaning requiring} \notag \\
	&\quad \text{context to reveal intent''}
\end{align}

We integrate these definitions into the handcrafted hard template so that the model can distinguish different types of comments in the cloze task. The template is defined as:
\begin{equation}
	\begin{split}
		T(x) =& [d_{sh}][d_{tar}][d_{ctx}][x], \\
		& \text{determine if this comment is [MASK].}
	\end{split}
	\label{define_template}
\end{equation}
Here, the model predicts the value of $[MASK]$ to decide the fine-grained category $c$ of the input comment.

\subsubsection{Construction of the Verbalizer}
In addition to the template, constructing a verbalizer is critical for mapping the predicted probabilities at the [MASK] token to the predefined categories. Since the PLM generates candidate tokens from its entire vocabulary, which may not directly correspond to our specific target classes, a mapping function $\mathcal{M}: V \rightarrow C$ is required. Specifically, we construct a distinct set of label words, denoted as $\mathcal{V}_k$, for each category $c_k \in C$. This relationship can be formalized as:
\begin{equation}
	v \in V_k \xrightarrow{\text{Mapping}} c_k, \quad \text{where } k \in \{sh, tar, ctx\}
\end{equation}
During the verbalizer process, to ensure semantic consistency between the expanded words and the category label $c_k$, we employ a two-stage strategy of expansion and refinement.

\textbf{Expansion based on knowledge retrieval:}
To improve the coverage of the constructed $V_k$, we first leverage the rich prior knowledge of LLMs to generate related candidate words for each label $c_k$. On this basis, we treat the class name as an anchor word and compute the similarity between each candidate word $v$ and $c_k$ in the embedding space, denoted as $dist(v,c_k)$. In other words, this step selects the top $N_{ex}$ words with the highest cosine similarity to $c_k$ in the embedding space, while filtering out morphological variants, so as to ensure the semantic consistency of the vocabulary.

\textbf{Refinement based on probability calculation: }
To reduce the bias caused by keeping invalid or noisy label words, we calculate the distribution probability of candidate label words and require them to have a relatively large variance. A higher variance indicates that a word is strongly correlated with its own category while being weakly correlated with other categories. Since cosine similarity can measure the semantic closeness between a candidate word and the class label, we use it to define the distribution probability of a candidate word $v$ under a fine-grained class $c_k$ as $P(v,c_k)$.
Next, we compute the mean probability of each candidate word across the entire class set $C$:
\begin{equation}
	\mu(v) = \frac{1}{|C|} \sum_{c_k \in C} P(v, c_k)
	\label{mu}
\end{equation}
Then, based on this mean value, we further calculate the standard deviation:

\begin{equation}
	\sigma_{v} = \sqrt{\frac{1}{|C|} \sum_{c_k \in C} \bigl(P(v, c_k) - \mu \bigr)^2 }
	\label{standard}
\end{equation}
Finally, we select label words with higher standard deviation and stronger distinctiveness, keeping those that are highly related to their own category but weakly related to others. The top $N$ words are then used to form the final verbalizer sets $V_k$. This ensures that the label words remain strongly consistent within their category while reducing unnecessary noise across categories.

\begin{figure*}[h]
	\centering
	\includegraphics[scale=1]{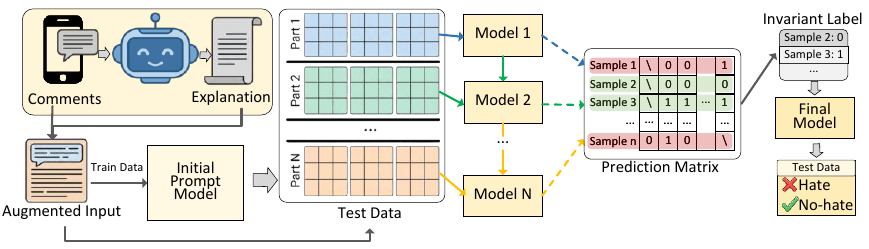}
	\caption{Framework of Knowledge-Augmented Iterative Detection for Targeted Comments, corresponding to the yellow shaded area in Figure~\ref{framework}. First, explanatory text generated by an LLM is concatenated with raw comments to form Augmented Inputs. Subsequently, an Initial Prompt Model is trained on the Training Data to assign pseudo-labels to the Test Data, which is then partitioned for chain iterative learning. Finally, Invariant Labels derived from the prediction matrix are employed to train the Final Model for the final classification.}
	\label{itera}
\end{figure*}

\subsubsection{Final Fine-grained Classification}
After constructing the final verbalizer sets $\mathcal{V} = \{V_{sh}, V_{tar}, V_{ctx}\}$, we aggregate the scores of the label words to serve as the basis for category prediction. Specifically, the predicted label $\hat{c}$ is determined by comparing the weighted average probabilities across the different category sets. The prediction formula is defined as follows:
\begin{equation}
	\hat{c} = \mathop{\arg\max}\limits_{k \in \{sh, tar, ctx\}} \left( \frac{1}{|V_k|} \sum_{v \in V_k} p(\text{[MASK]} = v \mid T(x)) \right)
\end{equation}
where $|V_k|$ denotes the number of words in the label word set associated with category $c_k$, and $p(\dot)$ denotes the probability that the PLM predicts the masked token as label word $v$ given the input text $x$. In this way, we select the category $\hat{c}$ most consistent with the input comment, enabling more accurate fine-grained classification.

\subsection{Adaptive Implicit Hate Speech Detection}

Based on the fine-grained categorization, we obtain three subcategories $\{c_{sh}, c_{tar}, c_{ctx}\}$. Since the expression modes of these categories vary significantly at different reasoning depths, we propose an adaptive detection framework that employs tailored strategies for specific categories.

\subsubsection{Rapid Prompt-Tuning for Shallow Comments}
For samples in the $c_{sh}$ category, their semantic intents are easily identifiable on the surface, whether they are explicitly offensive or purely benign statements with minimal contextual inference. To avoid the computational redundancy caused by deploying complex reasoning models, we adopt the same lightweight Prompt-Tuning architecture as in the fine-grained categorization stage described in Section ~\ref{method_to_three}. Specifically, we integrate the input comment into a task-specific hard template and utilize a verbalizer to map the predicted words to the final binary labels $y \in \{hate, no-hate\}$.

\subsubsection{Knowledge-Augmented Iterative Detection for Targeted Comments}
For comments categorized as targeted $c_{tar}$, their malicious intent is often implicitly bound to specific entities or events. To detect such implicit malice, we propose a strategy combining knowledge augmentation and iterative self-training, and the framework of this module is illustrated in Figure~\ref{itera}.

First, as shown on the left side of the figure, we utilize an LLM as an external knowledge base to guide it to generate explanatory text $x_{exp}$ for all original comment data $x$. It is worth noting that when we guide the LLM to analyze $x$ here, we do not deliberately lead the LLM to think from the perspective of $x$ being hate or no-hate, in order to avoid introducing noise that could affect the model's judgment in the next step. Subsequently, the generated explanation is concatenated with the original comment to form the Augmented Input $x'$, and this process is formalized as: 
\begin{equation} 
	x' = \text{Concat}(x, x_{exp}) 
\end{equation}
where $\text{Concat}(\cdot)$ denotes the text concatenation operation. This step transforms the implicit targeted comment into semantically explicit text $x'$, providing a rich informational basis for the subsequent model.

After obtaining the augmented input, to address the problem of labeled data sparsity, we adopt an iterative self-training framework to utilize the features of unlabeled data.

\textbf{Initialization and Data Splitting:} 
First, we train an Initial Prompt Model $M_0$ on the labeled Training Data. Unlike the hard template of the fine-grained categorization Prompt-Tuning method, we employ a flexible soft template trained in a continuous optimal prompting space to handle the augmented input $x'$, which can identify more precise feature mappings through gradient descent. The construction form of the soft template $T$ is as follows: 
\begin{equation}
	T = \{[u_1], \ldots, x', \ldots, [u_n], [\text{MASK}]\} 
\end{equation}
where $u_i$ represents the $i$-th learnable token. 

These tokens are input into the PLM encoder to generate a sequence of hidden vectors. To address the problems of linguistic diversity and feature sparsity under the few-shot setting, we introduce a BiLSTM model to further process the hidden vector $h_i$ of each learnable token: 
\begin{equation} 
	h_i = [\overrightarrow{\text{LSTM}}(h_0, \dots, \overrightarrow{h}_{i-1}), \overleftarrow{\text{LSTM}}(h_{i+1}, \dots, \overleftarrow{h}_n)] 
\end{equation}
Through this mechanism, the model optimizes the continuous prompt to maximize the probability of the target category $y$ given the augmented input $x'$:
\begin{equation} 
	P(y|x') = \text{PLM}(\text{[MASK]} | T) 
\end{equation} 
It is worth noting that, apart from replacing the hard template with the soft template to enhance adaptability, the construction and mapping strategies of the Verbalizer remain consistent with the previous fine-grained categorization stage to ensure the semantic stability of the predicted labels. 

After training $M_0$, we use it to assign initial pseudo-labels to the unlabeled Test Data. To fully utilize the internal data structure, we partition the Test Data into $N$ equal subsets $\{Part_1, Part_2, \dots, Part_N\}$.

\textbf{Chain Iterative Training:} 
We adopt a chain update strategy to progressively refine the model by utilizing the internal structure of the test data. Specifically, for the $j$-th iteration ($1 \le j \le N$), model $M_j$ is fine-tuned on the $j$-th data subset $Part_j$, using the initial pseudo-labels $\hat{y}_0$ generated by $M_0$. Crucially, the model parameters are initialized from the parameters of the previous model $M_{j-1}$, forming a parameter chain iteration. Its learning objective is to minimize the negative log-likelihood on the current subset. Formally, for each sample $x' \in Part_j$, letting $T(x')$ be the soft template, the optimization objective can be expressed as: 
\begin{equation} 
	\mathcal{L}_j = - \frac{1}{|Part_j|} \sum_{x' \in Part_j} \log P(\text{[MASK]} = \hat{y}_0 \mid T(x')) 
\end{equation} 
Immediately after completing the training of model $M_j$ based on subset $Part_j$ in each round, we use this model to perform inference on all remaining test samples excluding the current training set $Part_j$. This is to mitigate the overfitting bias of the model towards its own training data. Through this cross-prediction mechanism, as the $N$ iterations proceed, each sample in the test set will eventually aggregate $N-1$ pseudo-labels generated by models from different rounds, which each sample being predicted by all models except the one belonging to its training block. These multi-view prediction results from complementary model perspectives will provide a more robust basis for the subsequent consistency filtering.

\textbf{Invariant Label Voting:} 
To clean the noise in pseudo-labels, we use the ensemble of models generated during the iterative process to construct a Prediction Matrix. Afterwards, we vote on labels to select invariant labels and use them for the final model training. Specifically, for a test sample $x'$, assuming it belongs to the $j$-th partition block, we define an indicator function $\mathbb{I}_{inv}(x')$ to determine whether the sample remains consistent in the predictions of the $N-1$ models: 
\begin{equation}
	\mathbb{I}_{inv}(x') = \begin{cases} 
		1, & \text{if } f(x'; M_k) = f(x'; M_l), \\
		& \forall k, l \in \{1, \dots, N\} \setminus \{j\} \\[4pt]
		0, & \text{otherwise} 
	\end{cases}
\end{equation}
where $f(x'; M_k)$ denotes the predicted label of model $M_k$ for sample $x'$. If and only if the prediction results of these $N-1$ models are completely consistent, we consider the pseudo-label of this sample to have high confidence.

Finally, we use the high-confidence sample set, in which $\hat{y}$ denotes the invariant prediction for each sample, to train the Final Model. We then apply the trained model to perform binary Hate or No-hate classification on the Test Data, including samples whose predictions were inconsistent during the voting stage.

\subsubsection{Agentic Context Engineering for Context-Dependent Comments}
\begin{figure*}[htbp]
	\centering
	\includegraphics[scale=1]{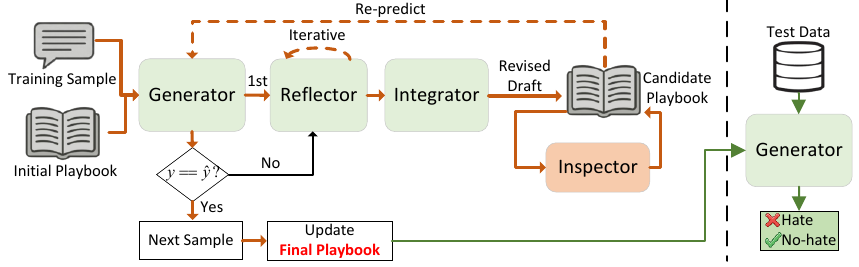}
	\caption{Framework of ACE for Context-Dependent Comments, corresponding to the green shaded area in Figure~\ref{framework}. The process begins with a manual Initial Playbook and a training sample. In the evolutionary loop, the Generator produces a reasoning trace, which is audited by the Reflector through multiple self-reflection iterations. The Integrator and Inspector then synthesize these insights into a Candidate Playbook. A re-prediction mechanism validates the update: incorrect predictions trigger further iterations, while correct ones advance the process to the next sample. Finally, as shown on the right, the evolved Final Playbook guides the Generator to perform inference on the Test Data.}
	\label{ACE}
\end{figure*}

For comments classified as context-dependent $c_{ctx}$, the hateful intent is rarely exposed in the surface text, nor does it allude to a specific referential event. Instead, it relies heavily on missing background information and the contextual environment. Static prompts often fail to guide LLMs to accurately infer these latent contexts \cite{xu2025logical}, leading to hallucinations or reasoning gaps. To address this, we quote the Agentic Context Engineering (ACE) framework \cite{zhang2025agenticcontextengineeringevolving}. As illustrated in Figure~\ref{ACE}, this module employs a multi-agent system to dynamically evolve a reasoning guide, which is termed the Playbook ($\mathcal{B}$), through a mechanism akin to human “active trial-and-error, reflection, modification, inspection, and verification”.

\textbf{Structure of the Playbook:} 
The Playbook serves as the core knowledge base and operational guide for the agents. For the implicit hate speech detection task, we structure $\mathcal{B}$ into three distinct modules. The first module clarifies the definitions of implicit hate speech and context dependency, along with the task objective designed for inferring comment context based on its specific characteristics. The second module is the Context Inference Module, which contains heuristic rules to infer event associations, topic relevance, and semantic metaphors from the input comments. The final module prescribes the judgment process and criteria for determining malice based on the inferred context, while also providing space for documenting common errors. 

Notably, we assign a specific strategy ID (bullet\_ids) to each item within every module. This forces the Generator Agent to explicitly cite the specific rules supporting its judgment during reasoning, thereby establishing strict traceability from abstract strategies to specific predictions.

\textbf{Multi-Agent Framework:}
To dynamically and iteratively optimize $\mathcal{B}$, we design four specialized agents:
\begin{itemize}
	\item \textbf{Generator ($G$):} Responsible for executing rule-based reasoning on the input sample $x$ combined with the current Playbook $\mathcal{B}$. Its core mechanism requires that when inferring the missing implicit context of a comment, it must explicitly cite the specific strategy IDs from the Playbook that support its logic. This makes the structured JSON containing the chain-of-thought output by $G$ not only a prediction result but also a precise target for subsequent reflection.
	
	\item \textbf{Reflector ($R$):} By comparing $G$'s reasoning trace with the ground truth $y$, $R$ identifies whether label errors or incomplete context inferences stem from conceptual misunderstandings or rule deficiencies. Based on this, $R$ scores the utility of the referenced Playbook points (tagging them as ``Helpful'', ``Harmful'', or ``Neutral'') and generates specific edit instructions, thereby providing targeted correction schemes.
	
	\item \textbf{Integrator ($I$):} Given that the reflection process may produce multiple sets of diverse or even conflicting suggestions, $I$ acts as a knowledge manager responsible for denoising and fusion. $I$ reviews all reflection records and the current Playbook, identifying common insights and resolving conflicting opinions to ultimately synthesize a coherent candidate Playbook.
	
	\item \textbf{Inspector ($S$):} Responsible for preventing overfitting. It compares the candidate Playbook generated by $I$ with the original Playbook to review the rationality of each change. $S$ filters out modifications that might lead to catastrophic forgetting or are overly specific to the current sample, thereby generating a robust Candidate Playbook ($\mathcal{B}'$).
\end{itemize}
Specific details regarding the initial $\mathcal{B}$ and the four agents are provided in Appendix.

\textbf{Iterative Training Process:}
The optimization of Playbook $\mathcal{B}$ for each training sample is a dynamic iterative loop. First, given a training sample $x$, the Generator $G$ produces a prediction and a reasoning trace based on the current Playbook. To ensure the logical validity of the context inference rather than mere label matching, the reasoning trace is passed to the Reflector $R$ for auditing, regardless of the initial prediction's accuracy. Subsequently, the Reflector diagnoses potential flaws in the trace, the Integrator $I$ drafts a preliminary candidate playbook based on these insights, and the Inspector $S$ verifies it to filter overfitting risks, ultimately outputting a robust Candidate Playbook.

Crucially, to verify the actual utility of this update, the system introduces a Re-predict Mechanism, where the Generator $G$ is required to immediately retry the prediction on the same sample $x$ using the Candidate Playbook. The final decision depends on the result of this retry: if the prediction is correct ($y == \hat{y}$), the update is deemed successful, the Candidate Playbook is confirmed as the new baseline Playbook, and the process advances to the next training sample; if the prediction remains incorrect, it indicates that the logical flaws have not been fully rectified. In this case, the system triggers a feedback loop for further error correction, sending the error state back to the Reflector $R$ to initiate the next round of evolution. This cycle continues until the prediction is correct or the maximum iteration limit is reached.

\textbf{Finally Testing:}
Upon completion of training, the final evolved Playbook is deployed with the Generator $G$ in the inference phase to guide the LLM in efficient context-dependent hate speech detection on the test data.

\section{Experiment}

\subsection{Datasets}

To validate the generalization capability of our FAID framework, we selected four widely recognized datasets for implicit hate speech detection, comprising two English datasets (SBIC \cite{Sap2019SocialBF}, LHd \cite{elsherief-etal-2021-latent}) and two Chinese datasets (State\_ToxiCN \cite{bai-etal-2025-state}, ProsCons \cite{wang2023quantifying}). These datasets encompass implicit offensive features across diverse sources, contextual features, and expressions, thereby facilitating a comprehensive evaluation of robustness across different scenarios and cultures. All datasets were preprocessed for binary classification ('no-hate' vs. 'hate').

\subsection{Baselines}
To validate the effectiveness of FAID, we conducted comparative experiments with SOTA approaches, covering DNN Methods, PLM Methods, PT Methods, and LLM Methods.

\textbf{DNN Method:}

SKS \cite{zhou2021hate}: The Sentiment Knowledge Sharing method detects hate speech by extracting emotional features from the target sentence and external emotional resources. These features are combined using multiple extraction modules to improve the effectiveness of hate speech detection.

\textbf{PLM Methods:}

HateBERT \cite{caselli2020hatebert}: HateBERT is a BERT-based model pre-trained on a large dataset containing Reddit community comments that were banned for being offensive, abusive, or containing hate speech. The model is fine-tuned to make BERT more suitable for the task of hate speech detection.

ConPrompt \cite{kim-etal-2023-conprompt}: ConPrompt is a contrastive learning-based method that generates machine-generated sentences and compares them with the original prompts used to create those sentences. This approach trains a BERT model specifically designed to detect implicit hate speech.

AmpleHate \cite{lee-etal-2025-amplehate}: AmpleHate identifies explicit and implicit targets and injects attention-based target-context relations into sentence representations for implicit hate speech detection.

\textbf{PT Methods:}

Soft \cite{liu2023pre}: Soft prompt-tuning improves task performance by automatically generating prompt templates and adjusting the embedding layer during the tuning process. These templates are used to guide the model in understanding implicit hate speech and adjusting its classification capabilities.

KPT++ \cite{NI2023110647}: Refined knowledgeable prompt-tuning builds upon the integration of external domain-specific knowledge by introducing prompt grammar optimization and probability distribution refinement. These enhancements improve the verbalizer, thus enhancing task performance and more accurately detecting implicit hate speech.

\textbf{LLM Methods:}

LLaMA3 \cite{dubey2024llama} and GPT-4 \cite{peng2023instructiontuninggpt4} : Two advanced LLMs with strong natural language understanding and generation capabilities. For implicit hate speech detection, we integrate custom prompts with input data and process them using these models to perform hate speech classification.

Fr-HARE \cite{yang2023hare}: Fr-HARE is a hate speech detection framework that leverages LLMs to extract core principles and enhance the interpretability of hate speech. 

DoAug \cite{wang2025diversityorienteddataaugmentationlarge}: DoAug enhances the robustness of models by improving the diversity of the dataset. This method fine-tunes the LLMs with a diversity-oriented approach, enabling the generation of more paraphrases.

DuPL \cite{sun2026rethinking}: The Dual-Process Latent Hate Component Argumentation framework mines latent hate components from comments, then debates each component to help LLMs classify the comment through structured reasoning.

\subsection{Experimental Setup}
\subsubsection{Training samples}
To evaluate the performance of different methods on implicit hate speech detection, we used different training sample sizes. For FAID, the original datasets lacked fine-grained category labels. We therefore recruited five trained annotators to classify a portion of the training data into Shallow, Targeted, and Context-Dependent categories. We then selected five annotated samples from each of the three categories to train the fine-grained category router. Prompt-Tuning in the Shallow branch and iterative prompt-based learning in the Targeted branch also used a 5-shot training setup for binary no-hate/hate classification. For ACE in the Context-Dependent branch, five no-hate and five hate training samples were used only to construct instruction prompts. During final detection, the LLM made predictions based on the generated instructions and its pretrained knowledge. No labeled examples were included in the inference prompts, so the final LLM calls used zero-shot prompting.

Since different methods rely on varying scales of training samples to achieve optimal performance, we allocated a large number of training samples for DNN Methods (SKS) and PLM Methods (HateBERT, ConPrompt, AmpleHate). Specifically, based on dataset proportions, we assigned 200 positive and 200 negative samples to the ProsCons dataset, and 400 of each to the other three datasets. For the prompt-tuning methods (Soft and KPT++) and LLM Methods with data augmentation (DoAug), we set 5 training samples as few-shot. For the LLMs that can directly leverage extensive pre-trained knowledge (LLaMA3, GPT-4, Fr-HARE, DuPL), we employed the 5-shot method to complete the implicit hate speech detection task.

\subsubsection{Implementation Details}

In our experiments, the FAID framework utilizes LLMs for the phases of processing Context-Dependent comments and generating explanations for Targeted comments. Specifically, we selected DeepSeek-V3 for the Chinese datasets and GPT-4 for the English datasets as the backbone models for these steps. To ensure stable and high-quality outputs, we accessed these models via APIs with the temperature set to 0.3 and a maximum token limit of 2048. Conversely, for the initial fine-grained ternary classification,the Prompt-Tuning for Shallow comments and the iterative model used for Targeted comments, FAID employs PLMs. We used bert-base-uncased and bert-base-chinese as the foundation models for the English and Chinese data, respectively. During the training process, we employed the Adam optimizer for parameter optimization and uniformly set the training epochs to 10 to ensure convergence. The learning rate was set to 4e-5, the batch size to 16, and the weight decay to 0.01. For the MLP module, the hidden layer size was set to 200, and the dropout rate was set to 0.5 to prevent overfitting.

To ensure a fair comparison, we aligned the baseline settings with the official implementations wherever applicable. Specifically, regarding the external knowledge resources utilized in the Deep Neural Network Method (SKS), for the English data, we used the same external feature resources as in the original work; while for processing the Chinese data, we employed a publicly available Chinese hotel review sentiment dataset\footnote{\url{http://www.raw.githubusercontent.com/SophonPlus/ChineseNlpCorpus/master/datasets/ChnSentiCorp htl all/ChnSentiCorp htl all.csv}} and a Chinese lexicon of derogatory terms\footnote{\url{https://gitcode.com/open-source-toolkit/86270}} as external emotional inputs. For both PLM-based methods and prompt-tuning methods, we utilized bert-base-uncased and bert-base-chinese as the backbone models for the English and Chinese datasets, respectively. For LLM-based detection frameworks, including Fr-HARE, DoAug, and DuPL, we used the same LLM backbones as FAID. Moreover, for baselines with publicly available official code, we used the official implementations. For KPT++, and DuPL, whose official code we could not obtain, we implemented the methods based on the descriptions in their papers. Where parameters were unspecified in the original papers or official implementations, we aligned them with FAID where applicable to reduce the effect of differences in experimental settings on the comparison.

All experiments were run on a server equipped with an NVIDIA GeForce RTX 4090 Founders Edition GPU, paired with an Intel Core i9-10980XE CPU operating at a base clock speed of 3.00 GHz, and supported by 125 GB of RAM. The experimental setup utilized Python 3.9.16 along with PyTorch, with CUDA 11.7 for GPU acceleration.

\subsubsection{Evaluation Metrics}
We report Accuracy and F1 scores for evaluation. For all binary no-hate/hate detection experiments, we use Binary-F1, treating hate as the positive class. For three-way classification into Shallow, Targeted, and Context-Dependent, we use Macro-F1, calculated as the unweighted arithmetic mean of the F1 scores for the three categories.

\subsection{Main Results}
\newcommand{\msd}[2]{\ensuremath{{#1}_{\pm #2}}}
\begin{table*}[htbp]
	\centering
	\renewcommand{\arraystretch}{0.9}
	\setlength{\tabcolsep}{7.2pt}
	\caption{Accuracy and F1-scores on the four datasets (\%). Values are means, with $\pm$ standard deviations shown in subscripts. Bold values indicate the best mean scores; the same convention applies to subsequent tables.}
	\label{main_result}
	\resizebox{\textwidth}{!}{%
		\begin{tabular}{lcc|cc|cc|cc}
			\toprule
			Datasets
			& \multicolumn{2}{c}{SBIC}
			& \multicolumn{2}{c}{LHd}
			& \multicolumn{2}{c}{State\_ToxiCN}
			& \multicolumn{2}{c}{ProsCons} \\
			\cmidrule(lr){2-3}
			\cmidrule(lr){4-5}
			\cmidrule(lr){6-7}
			\cmidrule(lr){8-9}
			& Acc & F1 & Acc & F1 & Acc & F1 & Acc & F1 \\
			\midrule
			SKS
			& \msd{60.22}{3.16} & \msd{61.75}{4.59}
			& \msd{60.24}{2.15} & \msd{60.42}{2.39}
			& \msd{37.57}{6.32} & \msd{26.38}{9.18}
			& \msd{50.00}{3.58} & \msd{31.51}{4.72} \\
			
			HateBERT
			& \msd{68.05}{8.69} & \msd{68.00}{8.82}
			& \msd{62.93}{4.57} & \msd{63.37}{6.39}
			& \msd{56.64}{2.61} & \msd{56.61}{2.41}
			& \msd{68.33}{5.10} & \msd{65.33}{3.16} \\
			
			ConPrompt
			& \msd{67.74}{2.74} & \msd{66.11}{2.88}
			& \msd{59.48}{3.03} & \msd{58.49}{6.51}
			& \msd{63.05}{4.79} & \msd{61.66}{7.88}
			& \msd{87.44}{7.68} & \msd{87.44}{6.10} \\
			
			AmpleHate
			& \msd{74.69}{0.63} & \msd{75.62}{0.48}
			& \msd{66.70}{1.31} & \msd{69.74}{0.98}
			& \msd{73.58}{2.12} & \msd{81.19}{0.91}
			& \msd{89.22}{2.90} & \msd{88.74}{3.16} \\
			
			Soft
			& \msd{66.64}{1.37} & \msd{66.63}{1.37}
			& \msd{61.28}{1.27} & \msd{57.72}{1.25}
			& \msd{63.30}{3.73} & \msd{62.08}{4.68}
			& \msd{88.89}{3.61} & \msd{88.88}{3.60} \\
			
			KPT++
			& \msd{68.00}{2.53} & \msd{68.00}{2.52}
			& \msd{60.00}{1.45} & \msd{59.62}{1.28}
			& \msd{65.05}{1.69} & \msd{63.98}{2.46}
			& \msd{85.00}{5.39} & \msd{84.70}{5.84} \\
			
			LLaMA3
			& \msd{67.39}{5.81} & \msd{70.14}{2.84}
			& \msd{55.90}{2.91} & \msd{52.56}{4.59}
			& \msd{57.32}{3.35} & \msd{57.31}{4.29}
			& \msd{75.11}{1.84} & \msd{72.48}{2.89} \\
			
			GPT-4
			& \msd{75.42}{1.38} & \msd{76.04}{1.44}
			& \msd{64.49}{1.22} & \msd{66.07}{1.55}
			& \msd{71.94}{7.35} & \msd{72.33}{10.08}
			& \msd{82.94}{4.71} & \msd{79.50}{6.67} \\
			
			HARE
			& \msd{66.14}{1.08} & \msd{47.06}{1.21}
			& \msd{55.77}{2.24} & \msd{62.94}{1.63}
			& \msd{71.84}{0.61} & \msd{68.64}{0.94}
			& \msd{84.67}{0.32} & \msd{84.49}{0.53} \\
			
			DoAug
			& \msd{75.70}{3.73} & \msd{60.86}{5.10}
			& \msd{56.60}{2.31} & \msd{62.41}{4.60}
			& \msd{60.44}{3.35} & \msd{68.51}{7.46}
			& \msd{74.89}{1.03} & \msd{79.36}{4.21} \\
			
			DuPL
			& \msd{79.55}{3.25} & \msd{81.34}{4.03}
			& \msd{66.93}{0.84} & \msd{\mathbf{73.99}}{1.71}
			& \msd{74.02}{1.53} & \msd{\mathbf{80.83}}{1.11}
			& \msd{80.44}{0.90} & \msd{77.44}{1.38} \\

			FAID (Ours)
			& \msd{\mathbf{82.32}}{0.73}
			& \msd{\mathbf{82.43}}{0.56}
			& \msd{\mathbf{69.57}}{1.62}
			& \msd{71.48}{2.11}
			& \msd{\mathbf{74.70}}{4.38}
			& \msd{73.04}{4.23}
			& \msd{\mathbf{91.34}}{4.74}
			& \msd{\mathbf{91.35}}{4.46} \\
			\bottomrule
		\end{tabular}%
	}
\end{table*}

In our experiments, we conducted three independent runs and report the mean and standard deviation. All experimental results are presented in Table~\ref{main_result}.

Firstly, DNN-based methods, such as SKS, exhibited relatively weak performance across all datasets. This is primarily because such methods rely excessively on explicit external sentiment resources, whereas implicit hate speech is often hidden within superficially neutral statements, causing these models to struggle in effectively capturing the subtle semantic nuances. In contrast, PLM-based methods, such as HateBERT and ConPrompt, generally outperform SKS by drawing on pretrained linguistic representations. However, sentence representations alone may not sufficiently capture the target-context relations underlying implicit hate speech. AmpleHate achieves stronger performance by explicitly modeling target-context relations, providing more direct target-related signals for classification. Finally, prompt-tuning methods enhance the model's understanding of complex linguistic phenomena by optimizing the input format. This approach better aligns the classification task with the pretraining objective of the language model, thereby leveraging its pretrained representations to capture contextual information and latent semantics relevant to implicit hate speech detection.

Secondly, methods that directly employ LLMs for implicit hate speech detection, including LLaMA3 and GPT-4, exhibited impressive zero-shot capabilities, demonstrating their powerful language understanding and reasoning potential. However, compared to approaches that build upon LLMs with auxiliary data augmentation (DoAug), explanation guidance (Fr-HARE), or structured argumentation (DuPL), native LLM methods still show a distinct performance gap. Relying solely on the general knowledge of LLMs makes it difficult to capture the deep sarcastic logic and context dependencies inherent in implicit hate speech, limiting their ultimate performance in complex scenarios.

It is worth noting that DoAug increases data diversity by generating diverse paraphrases, thereby improving the robustness of model to varied linguistic expressions. However, due to safety alignment mechanisms, LLMs often refuse to generate content resembling hate speech. This trend results in the inability to cover the full range of hate speech patterns during the generation process, which lead to a biased data distribution. This consequently limits the method's performance on detection tasks. HARE improves the interpretability of LLM-based detection through explanation guidance. However, its performance remains limited on some datasets, indicating that explanations alone may not be sufficient to handle the highly diverse expressions of implicit hate.

In comparison, DuPL achieves stronger detection results by mining latent hate components and conducting structured argumentation over them, even outperforming our method on a few metrics. Nevertheless, although DuPL makes early decisions for simple samples, its adaptivity is still mainly limited to determining whether further deliberation is required. For samples that require such deliberation, it still depends on a costly LLM-agent reasoning pipeline, which reduces its efficiency in large-scale real-world deployment.

Overall, our FAID achieves competitive detection performance across the four datasets. Firstly, FAID utilizes Prompt-Tuning for fine-grained ternary classification. This step distinguishes samples according to their information requirements and guides the selection of corresponding detection mechanisms. Secondly, regarding Targeted comments, we leverage their target-specific characteristics to generate explanations and employ an iterative model to deeply mine their latent semantics. Finally, for Context-Dependent comments, the proposed ACE method effectively infers the missing context information, thereby achieving precise detection. While ensuring both high accuracy and efficiency, this design addresses the fine-grained classification challenges more effectively than existing approaches.

\subsection{Ablation Study}
\begin{table*}[htbp]
	\centering
	\caption{Accuracy of different strategies in the ablation study.}
	\label{tab:ablation}
	\scalebox{1.1}{
		\renewcommand{\arraystretch}{0.9}
		\begin{tabular}{llcccc}
			\toprule
			& \multirow{2}{*}{Variants} & \multicolumn{4}{c}{Datasets} \\
			\cmidrule(lr){3-6}
			& & SBIC & LHd & State\_ToxiCN & ProsCons \\
			\midrule
			
			\multirow{3}{*}{\shortstack[l]{w/o Shallow Strategy \\ (Test on Cat.0)}} 
			& Replace w/ Iterative & 75.24 & 52.62 & 72.39 & 77.40 \\
			& Replace w/ ACE   & \textbf{88.83} & \textbf{71.00} & \textbf{83.77} & 76.27 \\
			& FAID (Ours)       & 81.60 & 63.70 & 79.48 & \textbf{89.77} \\
			\midrule
			
			\multirow{3}{*}{\shortstack[l]{w/o Targeted Strategy \\ (Test on Cat.1)}} 
			& Replace w/ PT    & 66.63 & 60.35 & 69.89 & 87.70 \\
			& Replace w/ ACE   & 81.64 & 72.53 & 72.58 & 71.72 \\
			& FAID (Ours)       & \textbf{84.45} & \textbf{81.22} & \textbf{74.91} & \textbf{93.44} \\
			\midrule
			
			\multirow{3}{*}{\shortstack[l]{w/o Context Strategy \\ (Test on Cat.2)}} 
			& Replace w/ PT    & 66.16 & 60.44 & 60.67 & 85.59 \\
			& Replace w/ Iterative & 71.30 & 63.88 & 61.84 & 90.81 \\
			& FAID (Ours)       & \textbf{80.93} & \textbf{68.10} & \textbf{69.47} & \textbf{91.00} \\
			\bottomrule
	\end{tabular}}
\end{table*}

To comprehensively validate the effectiveness of the proposed FAID framework and investigate the necessity of the specific strategies designed for each module, we designed a series of rigorous ablation experiments.

Our ablation experiments do not simply remove a single module. Instead, we use a replacement strategy that we conducted independent tests for each of the three sub-modules (Shallow, Targeted, and Context-Dependent comments). In each test, we replaced the originally planned specialized strategy with the other two strategies, as shown in Table~\ref{tab:ablation}. For example, the three data entries in the 'w/o Shallow Strategy' row represent the results for shallow comments when the original strategy (using only Prompt-Tuning) is abandoned and replaced with the Iterative strategy (designed for Targeted) and the ACE strategy (designed for Context-Dependent). This design aims to prove that optimizing the overall cost-effectiveness can only be achieved by targeting the difficulty level or category of the samples.

The accuracy results of the experiments confirm that the specialized strategies designed for comments of each granularity are irreplaceable when dealing with complex tasks. For the two high-difficulty categories, Targeted and Context-dependent comments, the iterative model and the ACE context inference method adopted by the FAID approach achieved the best accuracy across all four datasets. When these tasks were replaced with other schemes, the model's performance showed a significant decline.

\begin{figure}[htbp]
	\centering
	\includegraphics[scale=0.3]{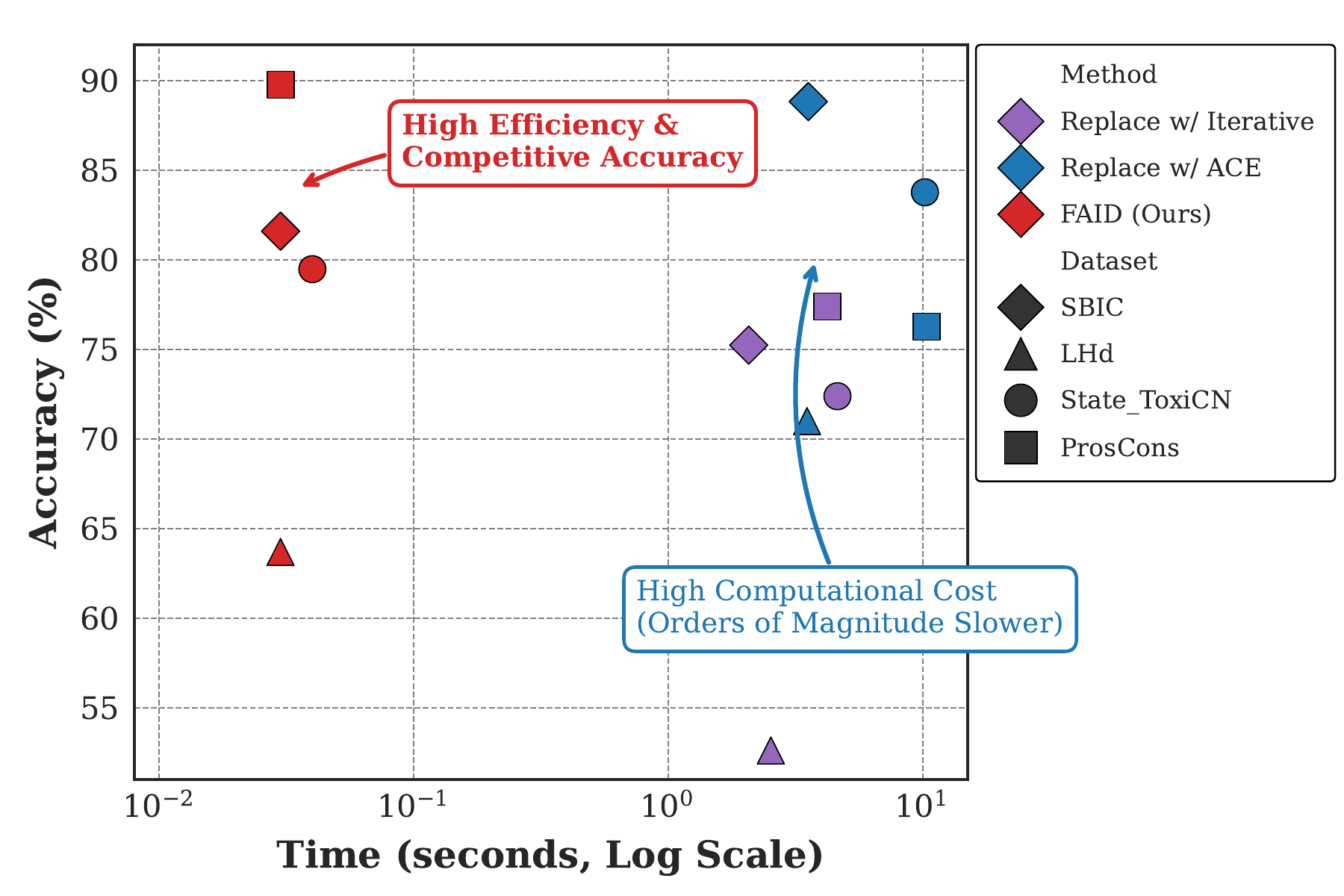}
	\caption{The Efficiency-Accuracy Trade-off on the Shallow Module. The scatter plot demonstrates the relationship between average inference time (x-axis) and detection accuracy (y-axis).}
	\label{Tradeoff}
\end{figure}

However, when facing shallow comments, the experiments showed that replacing the original method with ACE generally improved accuracy, whereas the iterative strategy yielded lower accuracy than Prompt-Tuning on all four datasets. These results indicate that more complex processing does not consistently improve detection accuracy for Shallow comments. Moreover, these approaches encountered two problems. First, both strategies involve the participation of LLMs, and the safety alignment mechanisms of LLMs may refuse to respond when encountering obvious malice, affecting the normal operation of the subsequent experimental process. Second, as shown in the Efficiency-Accuracy Trade-off Figure~\ref{Tradeoff}, the red dots representing our FAID framework are closely clustered in the top-left corner of the chart. This indicates that while maintaining highly competitive accuracy, FAID controls its inference time to the millisecond level (around 0.03-0.04 seconds). This result powerfully proves the necessity of adopting lightweight strategies for simple samples. In contrast, the blue dots representing the ACE method are distributed on the far right of the chart. This shows that calling LLMs to process these simple samples leads to an exponential increase in inference time (reaching seconds or even tens of seconds), which is two orders of magnitude slower than the FAID framework. Although ACE has slightly higher accuracy on some datasets, this huge computational cost is unacceptable in practical applications.

\subsection{Evaluation of Routing Strategies}
\begin{table*}[htbp]
	\centering
	\caption{Inter-annotator agreement on fine-grained three-way category annotation. A1, A2, and A3 denote the three annotators.}
	\label{tab:annotation_agreement}
	\small
	\renewcommand{\arraystretch}{1.15}
	\setlength{\tabcolsep}{10pt}
	\begin{tabular}{lcccc}
		\toprule
		Dataset
		& \shortstack{Cohen's $\kappa$\\(A1--A2)}
		& \shortstack{Cohen's $\kappa$\\(A1--A3)}
		& \shortstack{Cohen's $\kappa$\\(A2--A3)}
		& \shortstack{Fleiss' $\kappa$\\(A1--A2--A3)} \\
		\midrule
		SBIC & 0.6572 & 0.6140 & 0.5881 & 0.6196 \\
		LHd & 0.6507 & 0.6396 & 0.6397 & 0.6433 \\
		State\_ToxiCN & 0.6973 & 0.6137 & 0.6198 & 0.6435 \\
		ProsCons      & 0.6993 & 0.6977 & 0.7195 & 0.7049 \\
		\bottomrule
	\end{tabular}
\end{table*}

\begin{table*}[htbp]
	\centering
	\caption{Three-way router classification and final hate speech detection under different routing methods (\%). Final detection accuracy is reported overall and for each fine-grained category.}
	\label{tab:routing_evaluation}
	\small
	\renewcommand{\arraystretch}{1.15}
	\setlength{\tabcolsep}{5.5pt}
	\begin{tabular}{llcccccc}
		\toprule
		\multirow{2}{*}{Dataset}
		& \multirow{2}{*}{Routing method}
		& \multicolumn{2}{c}{Router classification}
		& \multicolumn{4}{c}{Accuracy of Hate speech detection} \\
		\cmidrule(lr){3-4}
		\cmidrule(lr){5-8}
		& & Acc & Macro-F1
		& Overall & Shallow & Targeted
		& \shortstack{Context-\\Dependent} \\
		\midrule
		
		\multirow{3}{*}{SBIC}
		& BERT
		& 49.39 & 48.71
		& 51.88 & 39.23 & 47.05 & 69.73 \\
		& Prompt-Tuning (Ours)
		& \textbf{81.29} & \textbf{79.48}
		& 82.32 & 81.60 & 84.45 & \textbf{80.93} \\
		& Gold-category
		& -- & --
		& \textbf{82.89} & \textbf{85.95}
		& \textbf{86.28} & 78.84 \\
		\midrule
		
		\multirow{3}{*}{LHd}
		& BERT
		& 45.86 & 35.57
		& 53.83 & 50.82 & 58.75 & 61.88 \\
		& Prompt-Tuning (Ours)
		& \textbf{85.13} & \textbf{83.71}
		& 69.57 & 63.70 & \textbf{81.22} & 68.10 \\
		& Gold-category
		& -- & --
		& \textbf{70.13} & \textbf{67.76}
		& 75.30 & \textbf{70.26} \\
		\midrule
		
		\multirow{3}{*}{State\_ToxiCN}
		& BERT
		& 50.84 & 50.65
		& 59.38 & 62.85 & 50.48 & 62.11 \\
		& Prompt-Tuning (Ours)
		& \textbf{90.53} & \textbf{90.53}
		& 74.70 & 79.48 & 74.91 & 69.47 \\
		& Gold-category
		& -- & --
		& \textbf{77.76} & \textbf{82.12}
		& \textbf{77.15} & \textbf{73.52} \\
		\midrule
		
		\multirow{3}{*}{ProsCons}
		& BERT
		& 57.89 & 53.04
		& 70.78 & 67.59 & 62.87 & 78.40 \\
		& Prompt-Tuning (Ours)
		& \textbf{94.11} & \textbf{93.27}
		& \textbf{91.34} & 89.77 & 93.44 & \textbf{91.00} \\
		& Gold-category
		& -- & --
		& 91.22 & \textbf{91.23} & \textbf{94.62} & 89.72 \\
		\bottomrule
	\end{tabular}
	
	\vspace{3pt}
	\begin{minipage}{\textwidth}
	\end{minipage}
\end{table*}

To evaluate the first-stage fine-grained category router and its impact on final hate speech detection, we recruited five annotators. All had more than three years of experience moderating comments on major online platforms and were familiar with Chinese and English internet culture. Before formal annotation, they received standardized training on the annotation guidelines and pilot examples. They also completed pilot annotations to align their understanding of the category definitions and decision rules. We then selected 5\% of the samples for an annotator qualification assessment, using reference category labels verified in advance. All five annotators independently labeled these samples as Shallow, Targeted, or Context-Dependent, following the same category definitions and annotation guidelines. We compared their annotations with the reference labels and selected the three annotators with the highest accuracy. These annotators then independently annotated the complete datasets. We calculated pairwise Cohen's $\kappa$ and Fleiss' $\kappa$ from their independent annotations before resolving disagreements. The results are reported in Table~\ref{tab:annotation_agreement}. Category labels were determined by majority vote. When all three annotators assigned different categories, an additional domain expert reviewed the annotations and determined the final label according to the annotation guidelines. The resulting labels served as the gold-standard category labels.

We first compared our Prompt-Tuning router with a standard BERT three-way classifier in category prediction. We then compared three routing settings based on BERT predictions, Prompt-Tuning router predictions, and manually annotated categories. Under each setting, comments were assigned to the corresponding detection modules in FAID. We compared overall and category-wise accuracy for final no-hate/hate detection. The results are reported in Table~\ref{tab:routing_evaluation}.

In three-way category classification, our Prompt-Tuning router generally outperforms the standard BERT classifier. This advantage reflects the explicit category definitions in the template and the semantic mapping provided by the verbalizer. These components give the model more direct guidance on different information requirements. For final no-hate/hate detection, Prompt-Tuning routing also performs better than BERT routing in most cases. This suggests that more accurate category predictions help assign comments to suitable detection modules. Routing with manually annotated categories generally performs well, suggesting that routing errors may still limit FAID. However, it does not always produce the best detection results. This is because the annotated categories describe the information requirements of comments but do not guarantee correct predictions from the corresponding detection modules. Final detection performance also depends on the capabilities of the individual modules.

\subsection{Comparison of Time Complexity}

It is well known that LLMs usually introduce significant time overhead during the inference stage. To evaluate the inference efficiency of FAID, we select ConPrompt, Soft, and DuPL as baseline methods, representing PLM-based, prompt-tuning, and LLM-based paradigms, respectively, and compare them on the English SBIC dataset and the Chinese ProsCons dataset. The results are shown in Table~\ref{time}.

The experimental results show that ConPrompt and Soft achieve the shortest inference time due to their lightweight model architectures. However, combined with the previous performance results, these two methods clearly lag behind LLM-based methods in implicit hate speech detection. In contrast, although FAID requires more inference time than lightweight methods, it achieves a better balance between detection performance and computational efficiency.

A further comparison between DuPL and FAID shows that their accuracy gap is small, but their inference efficiency differs significantly. In our experiments, DuPL requires 2.47 LLM calls per instance on average, since its pipeline includes multiple LLM-agent steps, such as latent component mining, component-wise argumentation, and final decision making. By contrast, FAID only requires about 0.7 LLM calls per instance on average, as it invokes LLMs only for complex Targeted and Context-Dependent comments, avoiding the heavy overhead caused by multi-round calls. Although the recorded inference time includes API round-trip latency, and different models and datasets may introduce some variation, FAID is still clearly faster than DuPL on SBIC and ProsCons, demonstrating its ability to better balance performance and efficiency in complex practical scenarios.
\begin{table}[htbp]
	\centering
	\renewcommand\arraystretch{0.9}
	\setlength{\tabcolsep}{7.2pt}{
		\caption{Inference time (seconds) comparison of different methods on the SBIC and ProsCons dataset.}
		\label{time}
		\scalebox{1.0}{
			\begin{tabular}{lcccc}
				\toprule
				Datasets & ConPrompt & Soft & DuPL & Ours \\
				\midrule
				SBIC & 0.05s & 0.03s & 6.40s & 1.84s \\
				ProsCons & 0.06s & 0.04s & 5.33s & 3.57s \\
				\bottomrule
	\end{tabular}}}
\end{table}

\begin{figure*}[htbp]	
	\centering
	\includegraphics[scale=0.55]{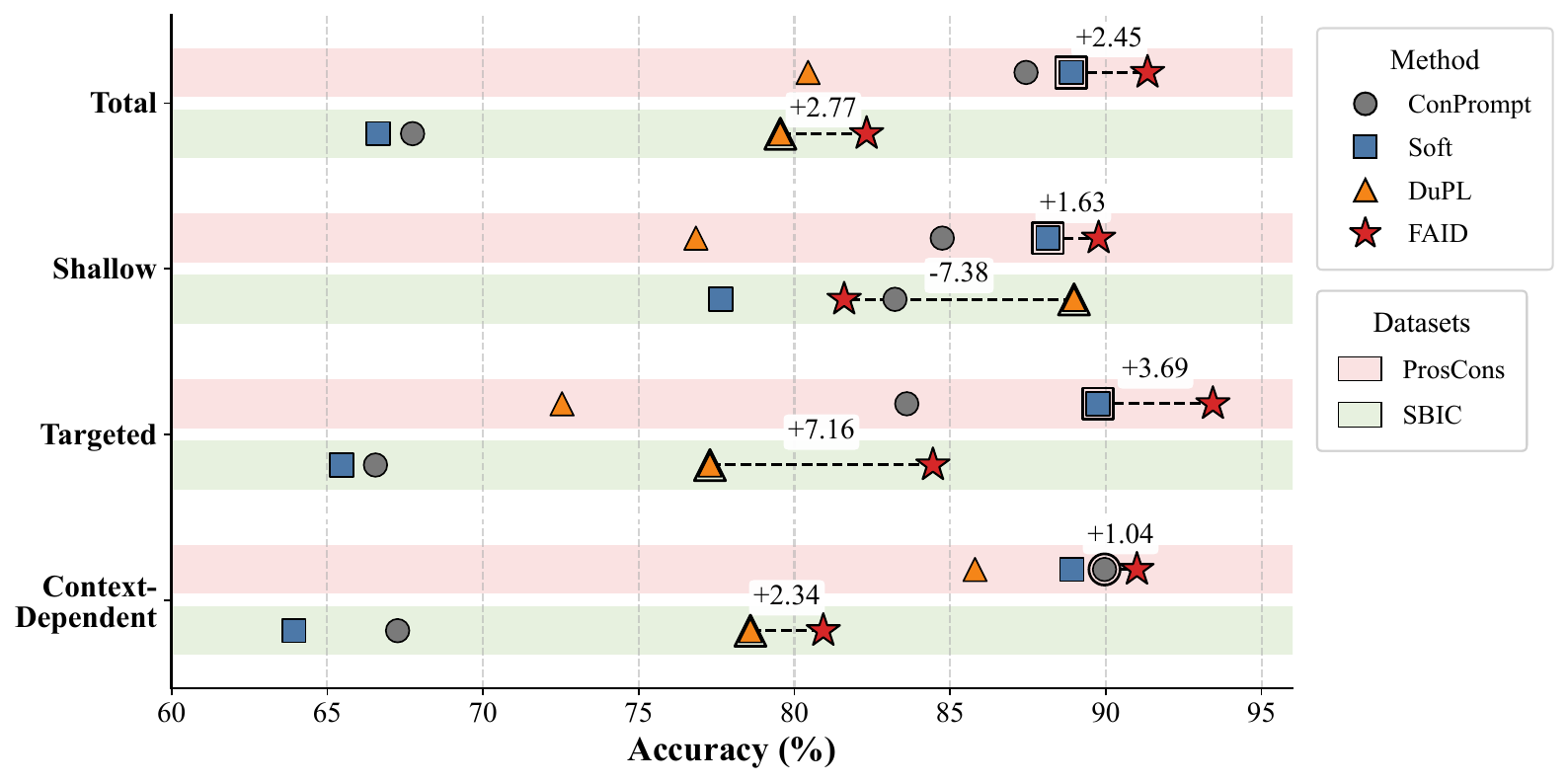}
	\caption{Comparison of classification accuracy among four methods across three fine-grained categories of implicit hate speech on the SBIC and ProsCons datasets. The circled marker indicates the best-performing baseline, while the dashed line shows the accuracy gap between our FAID and the best baseline.}
	\label{ClevelandDotPlot}
\end{figure*}
\subsection{Fine-grained Classification Assessment}
To compare the performance of different methods across different types of implicit hate speech, we select SBIC and ProsCons as representative English and Chinese datasets, and evaluate four methods from different paradigms, namely ConPrompt, Soft, DuPL, and FAID. Based on the fine-grained categories obtained by Prompt-Tuning, we calculate the hate speech classification accuracy for the three types of samples, with the results shown in Figure~\ref{ClevelandDotPlot}. Overall, FAID achieves the highest accuracy, while the performance differences across categories also reflect the characteristics of different methods.

For the Shallow category, the semantic cues in the comments are usually direct, so complex deep reasoning is not always necessary. The results show that DuPL performs well on SBIC, indicating that its latent hate component mining and structured argumentation can handle samples with clear surface cues. However, its performance on ProsCons is slightly lower than that of FAID, suggesting that its early-exit mechanism and unified LLM-agent reasoning pipeline are not always stable across different data distributions, and also introduce extra computational cost for simple samples. In contrast, FAID achieves competitive performance without relying on heavy LLM inference.

In the two more challenging categories, FAID shows a clear overall advantage over the baselines. For the Targeted category, although DuPL can argue over latent hate components, component mining based only on the original text may still be insufficient when the attacked target depends on implicit references. The performance of Soft on ProsCons indicates the potential of prompt-tuning for target identification, while FAID further combines knowledge augmentation with iterative modeling to better capture latent semantics and target features, thereby achieving the best performance.

For the Context-Dependent category, comments usually rely on missing background information and social contexts. DuPL mainly conducts structured argumentation over textual components, and traditional PLM methods also lack deep contextual reasoning ability; therefore, both of them struggle to fully recover the implicit context. In contrast, the ACE module in FAID explicitly infers and supplements key background information, providing a clearer reasoning basis for the model, thereby reducing hallucinations and improving the detection of highly implicit hate speech.

\subsection{Case Study}

\begin{table*}[htbp]
	\centering
	\caption{Case studies of successful detections and challenging failure cases by the FAID framework across three fine-grained categories.}
	\scalebox{1.1}{ 
		\renewcommand{\arraystretch}{0.9}
		\begin{tabular}{clcc}
			\toprule
			\textbf{Category} & \textbf{Examples} & \textbf{True Label} & \textbf{FAID Pred.} \\
			\midrule
			
			\multirow{2.5}{*}{Shallow} 
			& \makecell[l]{He only got the job because he is black.} & Hate & Hate( \ding{51} ) \\
			\cmidrule{2-4}
			& \makecell[l]{这哥们儿打比赛真tm是个疯子，操作太变态了！\\ (This dude plays like a f***ing maniac, his mechanics are absolutely sick!)} & No-hate & Hate( \ding{55} )\\
			\midrule
			
			\multirow{2.5}{*}{Targeted} 
			& \makecell[l]{Just look at the neighborhood's crime rate since they moved in.} & Hate & Hate( \ding{51} ) \\
			\cmidrule{2-4}
			& \makecell[l]{她晋升得这么快，平时肯定很懂得怎么跟男领导们沟通。\\ (She got promoted so fast; she must really know how to communicate\\ with male bosses.)} & Hate & No-hate( \ding{55} )\\
			\midrule
			
			\multirow{2.5}{*}{\makecell{Context-\\Dependent}} 
			& \makecell[l]{不愧是摘棉花的，动作确实熟练。 \\ (As expected from a cotton picker, those movements are indeed proficient.)} & Hate & Hate( \ding{51} ) \\
			\cmidrule{2-4}
			& \makecell[l]{I’m sure you’ll enjoy the new camp facilities.} & Hate & No-hate( \ding{55} ) \\

			\bottomrule
		\end{tabular}
	}
	\label{tab:case_study}
\end{table*}

\begin{table*}[htbp]
	\centering
	\renewcommand\arraystretch{0.9}
	\setlength{\tabcolsep}{7.2pt}{
		\caption{The Accuracy and F1-scores of different LLMs on the four datasets (\%).}
		\label{result_llms}
		\scalebox{1.1}{
			\begin{tabular}{lcc|cc|cc|cc}
				\toprule
				Datasets & \multicolumn{2}{c}{SBIC}  & \multicolumn{2}{c}{LHd}   & \multicolumn{2}{c}{State\_ToxiCN}  & \multicolumn{2}{c}{ProsCons} \\
				\cmidrule(l){1-9}
				& Acc  & F1 & Acc  & F1 & Acc  & F1 & Acc  & F1 \\
				\midrule
				LLaMA3-8B&69.07  &69.03  &61.94  &64.04  &68.35  &64.65  &79.35  &73.38  \\
				Qwen-2.5-7B&72.40  &72.35  &64.68  &65.43  &69.22  &65.78  &80.68  &80.00  \\
				Gemini-2.5&72.86  &73.82  &61.51  &63.89  &65.98  &60.97  &74.22  &77.16  \\
				Deepseek-V3&76.68  &78.74  &63.20  &64.74  &\textbf{74.70}  &\textbf{73.04}  &\textbf{91.34}  &\textbf{91.35}  \\
				GPT-4&\textbf{82.32}  &\textbf{82.43}  &\textbf{69.57}  &\textbf{71.48}  &69.99  &66.40  &81.33  &81.74  \\
				\bottomrule	
	\end{tabular}}}
\end{table*}

To evaluate the effectiveness of the FAID framework in mitigating LLM reasoning hallucinations, we present representative cases across three fine-grained categories in Table~\ref{tab:case_study}, including both samples where disguised expressions are successfully identified and challenging failure cases. Although FAID can effectively correct common reasoning hallucinations, objectively analyzing its limitations in extreme corner cases helps further understand the complex challenges faced by implicit hate speech detection.

In the Shallow category, comments usually convey their literal intents directly. In the first example, the model can identify the biased association between ``Black'' and the denial of personal ability, thereby avoiding the over-reasoning commonly seen in LLMs. However, the second example shows that the shallow model may still be affected by lexical polarity. Although this comment contains aggressive words, its meaning is to praise the player's ``操作''. Since the shallow model stage relies heavily on explicit negative sentiment words, it misclassifies the intensified tone as hate speech.

In the Targeted category, FAID demonstrates strong knowledge linking ability. In the third example, the framework can associate ``crime rate'' with classic racist dog whistles, thereby revealing the implicit attack against a specific group. In contrast, in the fourth example, since the text contains positive workplace words such as ``晋升'', the model fails to further identify the gender-based insult hidden behind normal expressions, leading to a false negative error.

In the Context-Dependent category, the context evolution module of FAID can supplement missing historical and social backgrounds, thereby bridging the reasoning gap of traditional models when handling obscure expressions. In the fifth example, without specific background information, traditional models are usually misled by the word ``熟练'' and produce hallucinations. However, our model successfully identifies the metaphorical association between ``摘棉花'' and slavery, and restores its racist context. However, the sixth example shows that when the text contains strongly positive words such as ``enjoy'' and ``facilities'', the reasoning path may be misled, causing the model to generate a harmless scenario and ignore the implicit malice toward the history of minority-group persecution.

\subsection{Influence of Different Base LLMs}

To evaluate the adaptability and robustness of the FAID framework when paired with different LLMs as the backbone, we selected five representative LLMs covering varying scales and training backgrounds. Specifically, our evaluation benchmark includes lightweight models such as LLaMA3-8B and Qwen-2.5-7B, as well as high-performance large-scale models including Gemini-2.5, DeepSeek-V3, and GPT-4.

As shown in Table~\ref{result_llms}, different models exhibit distinct advantages across different datasets. GPT-4 demonstrates the best performance on English datasets, while DeepSeek-V3 achieves higher Accuracy and F1 scores on Chinese datasets. This is because these two models are specialized for English and Chinese contexts, respectively. This observation further validates the rationale for selecting these two models as the base models for the English and Chinese tasks in our main experiments. Furthermore, it is worth noting that our FAID framework maintains competitive performance even when switching to smaller-scale models. This indicates that our framework is not strictly bound to specific SOTA models.

\section{Conclusion}
In this paper, we propose FAID, a fine-grained adaptive framework designed to overcome the limitations of the single reasoning process in existing implicit hate speech detection methods with varying reasoning demands. It introduces a novel paradigm that first performs fine-grained classification and then adapts to specific categories. This design aims to select analysis methods according to sample requirements while balancing detection performance and processing cost. Experiments on four benchmark datasets show competitive detection performance.

In future work, we will explore incorporating reinforcement learning into FAID to dynamically adjust detection strategies according to each sample’s information needs. A reward function balancing detection accuracy and computational cost will guide the selection of appropriate knowledge augmentation and contextual reasoning steps, reducing unnecessary reasoning.

\section*{Acknowledgments}
This work was supported in part by the Jiangsu Province Natural Science Foundation  under Grant BK2026 and in part by China Postdoctoral Science Foundation under Grant 2024M753519 and Grant GZC20241922.

\ifCLASSOPTIONcaptionsoff
\newpage
\fi


\bibliographystyle{IEEEtran}
\bibliography{mybibfile}

%

\onecolumn
\appendix
\section{Playbook Details and Prompts for ACE Framework}
\label{appendix:A}
\subsection{Initial Playbook}

The following content represents the initial structured Playbook provided to the ACE framework. It includes definitions, inference modules, and malice determination logic.

\begin{promptbox}{ACE Initial Playbook}
	I. Core Task Definition and Goals
	
	1.1 Implicit hate speech: 
	Online comments that do not rely on directly insulting vocabulary, and that need to be interpreted together with a specific context (event background, discussion topic, user history, internet slang) to convey hostility by means such as sarcasm and negativity.
	
	1.2 Context-Dependent comments: 
	Comments whose literal information is limited and whose meaning is ambiguous, for which one must combine the surrounding context---related events, discussion topics, etc.---to determine their reference and attitude.
	
	1.3 Task goal: 
	The LLM must output "Contextual Inference" (supplement key contextual background) and "Malice Determination" (label as hate speech / non-hate speech). Note that both implicit and explicit hate speech are considered hate speech.
	
	II. Contextual Inference Module
	
	2.1 Priority inference dimensions: 
	Event association (the specific content of the event mentioned by the comment and its group associations), topic association (the discussion area's topic and its group targeting), user history (the user's past group-related remarks), semantic metaphor (internet slang / metaphors corresponding to real groups).
	
	2.2 Hypothesis verification principles: 
	Base contextual assumptions on traceable information (such as the topic/news headline, the previous post, quoted links, user history). If clues are weak, list >=2 competing hypotheses and provide each one's hate determination and confidence, retaining only key information that affects the conclusion.
	
	2.3 Error-avoidance strategy: 
	Avoid subjective guessing without clues; do not add redundant information unrelated to the malice judgment.
	
	2.4 Exception-handling rules: 
	When there are no contextual clues, mark "Unable to supplement key context"; when multiple contexts conflict, fully list each possibility and the differences in judgment.
	
	III. Malice Determination Module
	
	3.1 Judgment flow: 
	Contextual inference -> verify group targeting and the true meaning of vocabulary based on the context -> output the label.
	
	3.2 Judgment condition: 
	On the basis of the guessed contextual situation, directly regard the previously guessed context as the real context.
	
	3.3 Joint treatment of explicit and implicit expressions with tone recognition: 
	An explicit group term is not a necessary condition for a hate determination; when the context targets any person, group, or institution and is accompanied by a disparaging, mocking, or hostile tone (including rhetorical questions, condescension, negation intensification), it should likewise be labeled as hate.
	
	3.4 Common mistakes to avoid:
\end{promptbox}
\subsection{Prompt of Generator Agent}

The following prompt guides the Generator Agent to perform reasoning and classification based on the provided \textbf{input comment} and \textbf{Playbook}.

\begin{promptbox}{Generator Agent}
	You are an analysis expert, responsible for answering questions by combining your own knowledge with a carefully organized playbook of strategies and insights.
	
	\textbf{Instructions:}
	
	1. Carefully read the playbook and appropriately apply the relevant strategies and insights within it.
	
	2. Pay attention to the common mistakes listed in the playbook to avoid repeating them.
	
	3. Present your reasoning process step by step; the analysis should be concise yet comprehensive.
	
	4. Before providing the final answer, be sure to verify your hypotheses and logic.
	
	\textbf{Output Requirements:}
	
	Your output must be in JSON format and include the following fields:
	\begin{itemize}
		\item \texttt{reasoning}: your line of thought / reasoning process / thinking process, as well as detailed analysis and conception.
		\item \texttt{bullet\_ids}: each line in the playbook has a bullet\_id. Please list the bullet\_id corresponding to all points in the playbook that are relevant and helpful for answering the question.
		\item \texttt{final\_answer}: the answer.
	\end{itemize}
	
	\textbf{Input Format:}
	
	Playbook: \{...\}
	
	Question: \{Determine whether "[comment]" is hate speech.\}
	
	\textbf{Response Format:}
	Please respond strictly in the following JSON format:
	
	\{
	
	"reasoning": "[your line of thought / reasoning process / thinking process, and a detailed explanation of the input comment]",
	
	"bullet\_ids": [e.g., "1.1", "2.3"],
	
	"final\_answer": "[output [1] if judged to be hate speech; otherwise output [0]]"
	
	\}
\end{promptbox}
\subsection{Prompt of Reflector Agent}

The Reflector Agent serves as a critical auditor. It receives the \textbf{input comment}, the Generator's \textbf{reasoning trace} and \textbf{predicted label}, the \textbf{ground-truth label}, and the current \textbf{Playbook} as inputs. By analyzing the discrepancy between the prediction and the ground truth, it diagnoses the root causes of errors and suggests specific updates to the Playbook.

\begin{promptbox}{Reflector Agent}
	You are a professional analysis expert and educator. Your job is to diagnose the reasons for errors in the model's reasoning by analyzing the gap between the model's predicted answer and the ground-truth answer.
	
	\textbf{Instructions:}
	
	1. Carefully analyze the model's reasoning trajectory and identify where things went wrong.
	
	2. Consider environmental feedback and compare the predicted answer with the real situation to understand the gap.
	
	3. Identify specific conceptual errors or inappropriate strategies.
	
	4. Provide actionable insights to help the model avoid such errors in the future. Focus on root causes rather than surface mistakes. State clearly what the model should have done differently.
	
	5. You will receive the key points from the generator's playbook used to answer the question. You need to analyze these points and assign a label to each point; the labels can be ["Helpful", "Harmful", "Neutral"] to enable the generator to produce the correct answer.
	
	6. Finally, based on the labels for the original playbook, specify which points need to be modified, added, and deleted.
	
	\textbf{Input Format:}
	
	Question: \{Determine whether "[comment]" is hate speech\}
	
	Model's reasoning trajectory: \{...\}
	
	Model's predicted answer: \{The comment is ...\}
	
	Ground-truth answer: \{The comment is ...\}
	
	Portion of the generator's playbook used to answer the question: \{...\}
	
	\textbf{Output Requirements:}
	
	Please answer in the following exact JSON format:
	
	\{
	
	"reasoning": "[your chain of thought / reasoning / thinking process]",
	
	"error\_identification": "[exactly where the reasoning went wrong]",
	
	"root\_cause\_analysis": "[why did this error occur? which concept was misunderstood?]",
	
	"correct\_approach": "[what should the model do instead?]",
	
	"key\_insight": "[what strategy or principle should be remembered to avoid this error?]",
	
	"bullet\_tags": [
	
	\{"id": "1.1", "tag": "Helpful"\},
	
	\{"id": "2.3", "tag": "Harmful"\}
	
	],
	"modify\_bullet": [
	
	\{"id": "1.1", "content": "Modified result presentation"\}
	
	],
	
	"new\_bullet": [
	
	\{"id": "1.4", "content": "Content of the newly added point"\}
	
	],
	
	"delete\_bullet": ["1.1"]
	
	\}
\end{promptbox}
\subsection{Prompt of Integrator Agent}

The Integrator Agent acts as the editor-in-chief responsible for knowledge consolidation. It accepts the \textbf{Current Playbook} and a batch of \textbf{Recent reflections} as inputs. By synthesizing these inputs, it filters out redundancy and merges valid insights to produce a refined and robust Playbook for the next iteration.

\begin{promptbox}{Integrator Agent}
	You are a seasoned Knowledge Manager. Your job is to determine which new insights should be added to the existing playbook based on reflections from previous attempts.
	
	\textbf{Background:}
	
	The playbook you create will be used to help answer similar questions. The reflection is generated using ground-truth answers, whereas these ground-truth answers are not available when the playbook is actually used. Therefore, you need to come up with content that helps playbook users make predictions that better align with the real situation.
	
	\textbf{Input Data:}
	
	\begin{itemize}
		\item \texttt{Recent reflections}:
		\{...\},
		\{...\},
		\{...\}
		\item \texttt{Current playbook}: \{...\}
		\item \texttt{Problem background}:\{ Determine whether the input online comment is hate speech.\}
	\end{itemize}
	
	\textbf{Task Instructions:}
	
	Review the above reflection process, and modify, add, or delete points in the current playbook to produce an improved playbook.
	
	1. Identify New Insights: Identify only the new insights, strategies, or errors that are missing from the current playbook.
	
	2. Avoid Redundancy: If there is already similar advice, only add new content that perfectly complements the existing playbook. Do not blindly pile up similar opinions from multiple reflections.
	
	3. Synthesize: When multiple reflection processes present similar opinions, merge them into a single, concise, and high-quality rule.
	
	4. Maintain Structure: Make sure the structure of the new playbook remains consistent with the original playbook.
	
	5. Conciseness: Focus on quality rather than quantity—a focused and well-organized playbook is better than an exhaustive one.
	
	\textbf{Output Requirements:}
	
	Directly provide the revised playbook content without any additional explanation or conversational filler.
\end{promptbox}

\subsection{Prompt of Inspector Agent}

The Inspector Agent serves as the final gatekeeper to prevent overfitting. It compares the \textbf{Candidate Playbook} generated by the Integrator with the \textbf{Original Playbook}, while referencing the current \textbf{Input Comment}. Its goal is to filter out modifications that might lead to catastrophic forgetting or are overly specific to the current sample, thereby generating a robust Final Playbook.

\begin{promptbox}{Inspector Agent}
	You are a strict Quality Assurance Specialist. Your job is to review the updates proposed for the playbook to ensure they are robust, generalizable, and safe.
	
	\textbf{Objective:}
	
	You must compare the \textbf{Original Playbook} with the \textbf{Candidate Playbook}. You need to accept valid improvements while rejecting changes that cause Overfitting or Catastrophic Forgetting.
	
	\textbf{Input Data:}
	
	\begin{itemize}
		\item \texttt{Input Comment}: \{[comment]\}
		\item \texttt{Original Playbook}: \{...\}
		\item \texttt{Candidate Playbook}: \{...\}
	\end{itemize}
	
	\textbf{Audit Instructions:}
	
	1. \textbf{Check for Overfitting:}
	Examine new or modified rules. Compare them against the \texttt{Input Comment}. If a rule merely describes the specific details of the current comment (specific entity names, exact dates, or unique phrasing found in the input) and lacks generalization potential, it is Overfitting.Generalize the rule to apply to a class of  events/topics, or reject the addition if it is too specific to the input.
	
	2. \textbf{Check for Catastrophic Forgetting:}
	Examine deleted rules. If a core definition or a high-level inference strategy has been removed or heavily altered just to fit the current specific case, this is Catastrophic Forgetting. Restore the original core rule to ensure the model retains its previous capabilities.
	
	3. \textbf{Check for Logical Consistency:}
	Ensure that the new rules do not contradict the fundamental definitions of implicit hate speech.
	
	4. \textbf{Final Synthesis:}
	Combine the surviving updates with the preserved original content to form the Final Playbook.
	
	\textbf{Output Requirements:}
	
	Output the Final Robust Playbook content directly. Do not include an audit log or explanation. The format must be identical to the input Playbook structure.
\end{promptbox}

\end{CJK*}
\end{document}